\documentclass[journal]{IEEEtran}
\usepackage{amsmath,amsfonts}
\usepackage{algorithmic}
\usepackage{algorithm}
\usepackage{array}
\usepackage[caption=false,font=normalsize,labelfont=sf,textfont=sf]{subfig}
\usepackage{textcomp}
\usepackage{stfloats}
\usepackage{url}
\usepackage{verbatim}
\usepackage{graphicx}
\usepackage{cite}
\usepackage{listings}
\usepackage{array}
\usepackage{makecell} 
\usepackage{booktabs}

\usepackage{multirow} 
\usepackage[dvipsnames]{xcolor}   
\usepackage{listings}             

\usepackage{xcolor}  

\lstdefinestyle{myverilog}{
	language=Verilog,                             
	backgroundcolor=\color{White},        
	basicstyle=\ttfamily\small,                   
	keywordstyle=[1]\color{Blue}\bfseries,        
	keywordstyle=[2]\color{Green}\bfseries,        
	keywordstyle=[3]\color{Orange}\bfseries,      
	commentstyle=\itshape\color{Green},           
	stringstyle=\color{Red},                      
	numbers=none,                                  
	showspaces=false,                              
	showstringspaces=false,                        
	captionpos=t,                                  
	frame=lrtb,                                    
	breaklines=true,                               
	columns=fullflexible,                          
	morekeywords=[1]{include,module,inout,electrical,analog,begin,end,if,else,\$abstime}, 
	morekeywords=[2]{V, I, NN, GRU_NN, Readout_layer},                      
	morekeywords=[3]{Vams,timescale, disciplines, CDE, h, gnd, n, p},                 
}

\lstdefinestyle{python}{
	language=Python,                              
	backgroundcolor=\color{White},               
	basicstyle=\ttfamily\small,                  
	keywordstyle=[1]\color{Blue}\bfseries,       
	keywordstyle=[2]\color{Green}\bfseries,      
	keywordstyle=[3]\color{Orange}\bfseries,     
	commentstyle=\itshape\color{Gray},           
	stringstyle=\color{Red},                     
	numbers=none,                                
	numberstyle=\tiny\color{Gray},               
	showspaces=false,                            
	showstringspaces=false,                      
	captionpos=t,                                
	frame=lrtb,                                  
	breaklines=true,                             
	columns=fullflexible,                        
	morekeywords=[1]{import, def, lambda},       
	morekeywords=[2]{model, weight, bias},       
	morekeywords=[3]{Symbol, Matrix, Max},       
}
\usepackage{listings}
\usepackage{xcolor}

\begin{document}

\title{Hybrid Time-Domain Behavior Model Based on Neural Differential Equations and RNNs
}

\author{Zenghui Chang,Yang Zhang, Hu Tan,Hong Cai Chen,~\IEEEmembership{Member,~IEEE,}
\thanks{This paper was produced by the IEEE Publication Technology Group. They are in Piscataway, NJ.}
\thanks{}}

\markboth{Journal of \LaTeX\ Class Files,~Vol.~14, No.~8, August~2021}%
{Shell \MakeLowercase{\textit{et al.}}: A Sample Article Using IEEEtran.cls for IEEE Journals}

 
\maketitle

\begin{abstract}
Nonlinear dynamics system identification is crucial for circuit emulation. Traditional continuous-time domain modeling approaches have limitations in fitting capability and computational efficiency when used for modeling circuit IPs and device behaviors.
This paper presents a novel continuous-time domain hybrid modeling paradigm. It integrates neural network differential models with recurrent neural networks (RNNs), creating NODE-RNN and NCDE-RNN models based on neural ordinary differential equations (NODE) and neural controlled differential equations (NCDE), respectively.
Moreover, a hybrid adjoint method is designed for model training of this hybrid dynamic model.
Validation using real data from PIN diodes in high-power microwave environments shows NCDE-RNN improves fitting accuracy by 33\% over traditional NCDE, and NODE-RNN by 24\% over CTRNN, especially in capturing nonlinear memory effects.
The model has been successfully deployed in Verilog-A and validated through circuit emulation, confirming its compatibility with existing platforms and practical value.
This hybrid dynamics paradigm, by restructuring the neural differential equation solution path, offers new ideas for high-precision circuit time-domain modeling and is significant for complex nonlinear circuit system modeling.

\end{abstract}

\begin{IEEEkeywords}
	Transient simulation model, neural controlled differential equations, continuous-time neural networks, Verilog-A

\end{IEEEkeywords}

\section{Introduction}
Neural network - based transient behavior modeling of circuit IPs and devices has been successfully applied in various scenarios, including analog RF amplifiers \cite{ref1,ref2,ref3,ref4}, input - output drivers \cite{ref3,ref5,ref6}, combinational logic gates \cite{ref6,ref7}, and electrostatic discharge protection circuits \cite{ref8}. These examples often involve time - domain behavior modeling methods such as memory polynomials and other Volterra series \cite{ref12}. In the field of frequency - domain behavior modeling, algorithms widely used for large - signal nonlinear modeling include multi - harmonic distortion \cite{ref13}, X - parameter models \cite{ref10}, and Cardiff models \cite{ref11}. Compared with large - signal frequency - domain modeling, transient time - domain modeling has significant advantages in capturing complex nonlinear dynamics, handling non - stationary input signals, and directly obtaining transient responses.

Time - domain behavior modeling of devices and circuit IPs is similar to nonlinear system identification in control algorithm fields \cite{ref9,ref14,ref15}. Neural networks provide an effective tool for handling complex dynamic systems, including time - delay recurrent neural networks (TDRNN) \cite{ref1,ref7,ref8}, nonlinear autoregressive networks with exogenous inputs (NARX) \cite{ref2,ref3,ref4}, and continuous - time neural networks based on state - space models (ISSNN) \cite{ref6,ref16,ref28,ref29}. Continuous - time modeling methods have unique advantages in handling dynamic systems. They can naturally capture the continuity of system characteristics, providing high - precision system descriptions. These modeling methods show stronger robustness in system control and prediction, especially suitable for analyzing high - frequency or fast - changing systems. These three types of neural network behavior models can be implemented in widely - used device modeling languages such as Verilog - A \cite{ref17} \cite{ref18,ref19,ref16}, or deployed as sub - circuit structures in SPICE netlists \cite{ref6,ref20} for circuit - level transient simulation analysis.

However, discrete TDRNN needs to access multiple historical input steps during each calculation, significantly increasing simulation costs. Also, discrete - time modeling usually adopts fixed time steps, conflicting with the variable time - step flexibility of mainstream SPICE simulators \cite{ref21}, causing serious simulation accuracy and convergence issues. NARX models use historical outputs as current inputs, easily leading to instability in simulations \cite{ref16,ref22} and potential simulation failures. In contrast, the applicability of continuous - time recurrent neural networks for circuit simulation stability has been fully verified \cite{ref6,ref16}.

This paper explores the fusion of continuous - time - domain neural modeling and discrete recurrent structures, proposing two hybrid dynamics models for circuit behavior modeling: Neural Ordinary Differential Equation - RNN (NODE - RNN) and Neural Controlled Differential Equation - RNN (NCDE - RNN). Although the concept of NODE - RNN originates from \cite{ref41}, this study first applies it to circuit fields and implements it via Verilog - A for circuit time - domain modeling. More innovatively, this paper first proposes the NCDE - RNN model and proves its significantly better performance than the original NCDE architecture. The core mechanism of these two models lies in using neural differential equations to depict the continuous evolution of hidden states and introducing RNN to update states at discrete observation points. This effectively combines the advantages of continuous dynamics and discrete event learning. Through comparative experiments with traditional CTRNN and original NCDE, we verify the significant advantages of this hybrid modeling paradigm in capturing circuit nonlinear behaviors, especially in handling high - frequency transients and complex memory effects.

The remaining content of this paper is arranged as follows: Section II first reviews the application status of existing CTRNN models in circuit modeling, delves into the mathematical principles of NODE and NCDE and their physical interpretations in circuit models, then details the proposed hybrid dynamics model architecture and its mathematical superiority; Section III evaluates the modeling accuracy and fitting ability of different models through TCAD simulated data of PIN diodes in high - power microwave environments; Section IV focuses on the implementation of the model in Verilog - A, analyzing simulation performance indicators and discussing its engineering practicality; Section V summarizes the contributions and significance of this research and prospects future research directions.The VI section is dedicated to the discussion of the hybrid adjoint method proposed in this paper.

\section{Continuous-Time Domain Modeling Algorithms for Circuit Emulation}

\subsection{Continuous Recurrent Neural Network}
In continuous dynamic time-domain algorithms for circuit emulation, the Continuous-Time Recurrent Neural Network (CTRNN) \cite{ref16} is commonly used. Its mathematical expression is shown in equation (1).
\begin{align}
	\dot{\boldsymbol{x}(t)} = -\frac{1}{\tau}\boldsymbol{x}(t) + \sigma(\boldsymbol{A}\boldsymbol{x}(t) + \boldsymbol{B}\boldsymbol{u}(t) + \boldsymbol{b}_{u})
\end{align}
where $\boldsymbol{x}(t) \in \mathbb{R}^{n}$ is the state vector, $\boldsymbol{u}(t) \in \mathbb{R}^{m}$ is the input vector; $\boldsymbol{A} \in \mathbb{R}^{n\times n}$, $\boldsymbol{B} \in \mathbb{R}^{n\times m}$ are weight matrices, $\boldsymbol{b}_{u} \in \mathbb{R}^{n}$ is the bias vector, and $\tau$ is a scalar parameter. $\sigma$ is an element-wise nonlinear function, typically the hyperbolic tangent function. This model can be seen as the continuous-time counterpart of discrete-time networks. It models dynamic systems by considering the continuous change of the state vector $\boldsymbol{x}(t)$ over time and the impact of the input $\boldsymbol{u}(t)$ on the state.

\subsection{Neural Ordinary Differential Equations and Neural Controlled Differential Equations}

References \cite{ref16} and \cite{ref47} detail the use of Neural Ordinary Differential Equations (NODE) for modeling the time-domain behavior of circuit IPs. They show that this NODE model significantly reduces simulation time in verilog-A and lowers computational costs in power circuit modeling. In contrast, the Neural Controlled Differential Equations (NCDE) from \cite{ref38} have yet to be applied in this context.

\subsubsection{Neural Ordinary Differential Equations}

Neural Ordinary Differential Equations \cite{ref39} form a class of continuous-time models. They define the hidden state $\boldsymbol{x}(t)$ as the solution to an Initial Value Problem (IVP) of an ordinary differential equation, as shown in equation (2):
\begin{align}
	\frac{d\boldsymbol{x}(t)}{dt} = \boldsymbol{f}_\theta(\boldsymbol{x}(t),t) \quad \text{where} \quad \boldsymbol{x}(t_0) = \boldsymbol{x}_0
\end{align}

Here, the function $\boldsymbol{f}_\theta$ is specified by a neural network with parameters $\theta$, determining the dynamics of the hidden state. The hidden state $\boldsymbol{x}(t)$ is defined for all time points and can be evaluated at any required time using a numerical ODE solver:
\begin{align}
	\boldsymbol{x}_0,\ldots,\boldsymbol{x}_N = \text{ODESolve}(\boldsymbol{f}_\theta, \boldsymbol{x}_0, (t_0,\ldots,t_N))
\end{align}
Reference \cite{ref39} also demonstrates how to compute memory-efficient gradients related to parameters $\theta$ using adjoint sensitivity, and highlights the use of black-box ODE solvers as building blocks in larger models.

\subsubsection{Neural Controlled Differential Equations}

A limitation of Neural Ordinary Differential Equations is that $\boldsymbol{x}(t_1)$ is determined solely by $\boldsymbol{x}(t_0)$ and $\boldsymbol{f}_\theta$, which reduces the representation learning ability of NODE. Basic NODE may only be suitable for modeling circuits like oscillators that have no inputs. In most cases, we need to map inputs to outputs, similar to nonlinear dynamical system identification. To address this, Neural Controlled Differential Equations \cite{ref38} create an additional path $\boldsymbol{U}(t)$ from underlying time-series data, making $\boldsymbol{x}(t_1)$ depend on both $\boldsymbol{x}(t_0)$ and $\boldsymbol{U}(t)$. The initial value problem for NCDEs can be written as:

\begin{align}
\boldsymbol{x}(t_1) = \boldsymbol{x}(t_0) + \int_{\boldsymbol{t_0}}^{\boldsymbol{t_1}} \boldsymbol{f}(\boldsymbol{x}(t); \boldsymbol{f}_\theta) d\boldsymbol{U}(t) 
\end{align}

Or in the form as (5):

\begin{align}
\boldsymbol{x}(t_1) = \boldsymbol{x}(t_0) + \int_{\boldsymbol{t_0}}^{\boldsymbol{t_1}} \boldsymbol{f}(\boldsymbol{x}(t); \boldsymbol{f}_\theta) \frac{d\boldsymbol{U}(t)}{d\boldsymbol{t}} d\boldsymbol{t} 
\end{align}

Here, $\boldsymbol{U}(t)$ is the natural cubic spline path of the underlying time-series data. Specifically, this integral problem is known as a Riemann-Stieltjes integral (while NODEs use Riemann integrals). Although $\boldsymbol{U}(t)$ can be generated in other ways, the original authors of NCDEs recommend the natural cubic spline method due to its properties: 1) It is twice differentiable; 2) It has low computational cost; 3) The interpolated $\boldsymbol{U}(t)$ is continuous with respect to time $t$. This paper also uses the natural cubic spline method during training.

\subsubsection{Connection between Neural Controlled Differential Equations and the Current Continuity Equation}
Neural Controlled Differential Equations encode the voltage dynamics $\frac{dV}{dt}$ as the Riemann-Stieltjes integral kernel of the control path $\boldsymbol{U}(t)$ through diffeomorphic mapping. This establishes a direct mathematical isomorphism with the current equation $I(t) = \frac{dQ}{dt} = C\frac{dV}{dt}$. Let $\boldsymbol{U}(t)$ be constructed from the voltage signal as $\boldsymbol{U}(t) = \int_{t_0}^t \frac{dV(\tau)}{d\tau} d\tau$, then the state evolution of NCDE can be expressed as:
\begin{align}
	\frac{d\boldsymbol{x}(t)}{dt} = \boldsymbol{f}_\theta\big(\boldsymbol{x}(t)\big) \frac{d\boldsymbol{U}(t)}{dt} = \boldsymbol{f}_\theta(\boldsymbol{x}(t)) \frac{dV(t)}{dt}
\end{align}
Its implicit solution is $\boldsymbol{x}(t) = \boldsymbol{x}(t_0) + \int_{t_0}^t \boldsymbol{f}_\theta(\boldsymbol{x}(\tau)) \frac{dV(\tau)}{d\tau} d\tau$. Specifically, when the output layer is defined as $I(t) = \boldsymbol{g}_\phi(\boldsymbol{x}(t))$ and $\boldsymbol{f}_\theta(\cdot)$ learns the capacitance coefficient $C$, the network directly implements the physical constraint $I = C\frac{dV}{dt}$ through the path integral of $\frac{dV}{dt}$. The adjoint gradient $\frac{\partial I}{\partial C} = \mathbb{E}_t[\frac{dV}{dt}]$ can be analytically computed. This structure seamlessly embeds the differential-type constitutive equations of circuit models into the control terms of neural differential equations, making it mathematically suitable for time-domain behavior modeling of circuit IPs or devices.

\subsection{Hybrid Model Introduced in This Paper for Circuit Modeling}
This paper presents a hybrid dynamics model that incorporates discrete RNN - based dynamical modeling into the solution path of neural network continuous differential equations. Experiments show this improvement enhances modeling capabilities, particularly for nonlinear systems. The flowcharts for NODE - RNN and NCDE - RNN are shown in Figure 1.

\begin{figure*}[t]
	\centering
	\includegraphics[width=0.80\textwidth]{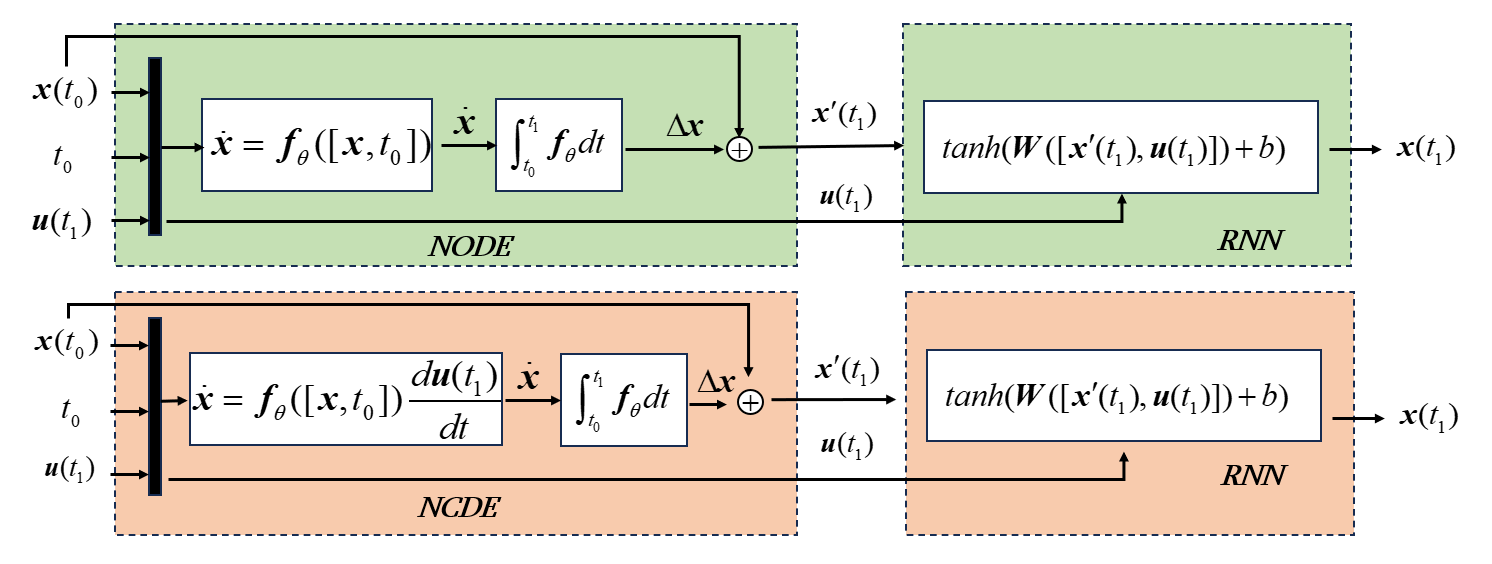}
	\caption{The solution procedures of NODE-RNN and NCDE-RNN }
	\label{fig:screenshot012}
\end{figure*}

\subsubsection{NODE-RNN}

The hybrid dynamics model of NODE-RNN was first proposed in \cite{ref41}. The core idea is to model the hidden states in the continuous time domain using neural ordinary differential equations and update the hidden states with a standard RNN at each observation. The algorithmic process of NODE-RNN is shown in Pseudo - code 1.

\begin{algorithm}
	\caption{The NODE-RNN}\label{alg:ode_gru}
	\begin{algorithmic}[1] 
		\STATE \textbf{Input:} Data points and their timestamps $\{\boldsymbol{u}_i, \boldsymbol{t}_i\}_{i=1..N}$
		\STATE $\boldsymbol{x}_0 \gets 0$
		\FOR{$i$ in $1, 2, \ldots, N$}
		\STATE $\boldsymbol{x}_i' \gets \boldsymbol{ODESolve}(\boldsymbol{f}_\theta, \boldsymbol{x}_{i-1}, (\boldsymbol{t}_{i-1}, \boldsymbol{t}_i))$ 
		\STATE $\boldsymbol{x}_i \gets \boldsymbol{RNN}(\boldsymbol{x}_i', \boldsymbol{u}_i)$ 
		\ENDFOR
		\STATE $\boldsymbol{o}_i \gets \boldsymbol{OutputNN}(\boldsymbol{x}_i)$ \textbf{for all} $i = 1..N$
		\STATE \textbf{Return:} $\{\boldsymbol{o}_i\}_{i=1..N}; \mathbf{x}_N$
	\end{algorithmic}
\end{algorithm}

\subsubsection{NCDE-RNN Model Proposed in This Paper}
Drawing on the modification approach of NODE-RNN to ODE models, this paper first proposes a hybrid model combining Neural Controlled Differential Equations with RNN. Specifically, an RNN - based observation network with discrete - event learning ability is added at each controlled integration step. The algorithm pseudocode is shown in Algorithm 2. Experiments prove this hybrid model has higher modeling capability than the original NCDE model. It offers a new approach for complex time - domain behavior modeling and further demonstrates the hybrid model's advantages.

\begin{algorithm}
	\caption{The NCDE-RNN}
	\begin{algorithmic}[1] 
		\STATE \textbf{Input:} Data points and their timestamps $\{\boldsymbol{u}_i, \boldsymbol{t}_i\}_{i=1..N}$
		\STATE $\boldsymbol{x}_0 \gets 0$
		\FOR{$i$ in $1, 2, \ldots, N$}
		\STATE $ \boldsymbol{x}_i' = \boldsymbol{x}_{i-1} + \int_{\boldsymbol{t_{i-1}}}^{\boldsymbol{t_{i}}} \boldsymbol{f}(\boldsymbol{x}(t); \boldsymbol{f}_\theta) \frac{d\boldsymbol{u}(t)}{d\boldsymbol{t}} d\boldsymbol{t}$
		\STATE $\boldsymbol{x}_i \gets \boldsymbol{RNN}(\boldsymbol{x}_i', \boldsymbol{u}_i)$ 
		\ENDFOR
		\STATE $\boldsymbol{o}_i \gets \boldsymbol{OutputNN}(\boldsymbol{x}_i)$ \textbf{for all} $i = 1..N$
		\STATE \textbf{Return:} $\{\boldsymbol{o}_i\}_{i=1..N}; \mathbf{x}_N$
	\end{algorithmic}
\end{algorithm}

\subsection{Improvement of the Readout Layer}
The readout layer of CTRNN or NODE is typically set as ${\boldsymbol{y}(t)}$ equal to $\mathbf{NN}{(\boldsymbol{x}_t^0, \ldots, \boldsymbol{x}_t^n)}$, where $\mathbf{NN}$ represents a neural network, usually a fully connected layer. Inspired by the controlled model (11), the output is also directly influenced by the input. We introduce the original input into the readout layer and use a two - layer fully connected layer with a \texttt{Tanh()} non - linear activation function in between. Experiments show this improvement greatly enhances modeling capability. Thus, in the four models tested in this paper - NCDE, NCDE - RNN, CTRNN, and NODE - RNN - this readout layer is adopted. Moreover, since the input derivative is obtainable, the two NCDE models incorporate the first - order derivative of the input $\dot{\boldsymbol{u}}(t)$ into the readout layer calculation.

\begin{align}
\begin{cases}
	\dot{\boldsymbol{x}}(t) = \boldsymbol{A} \boldsymbol{x}(t) + \boldsymbol{B} \boldsymbol{u}(t), \\
	\boldsymbol{y}(t) = \boldsymbol{C} \boldsymbol{x}(t) + \boldsymbol{D} \boldsymbol{u}(t) 
\end{cases}
\end{align}

\begin{align}
\boldsymbol{y}(t) = {\boldsymbol{W}_2}({\rm{tanh(}}{\boldsymbol{W}_1}([{\boldsymbol{x}_t};\boldsymbol{u}(t)]) + {b_1}) + {b_2}
\end{align}

\section{Experimental Evaluation}

\subsection{Dataset Acquisition and Model Training Validation}
To assess the performance of different models in nonlinear device time - domain modeling, this paper focuses on modeling the voltage excitation and response waveforms of PIN diodes in high - power microwave environments. To obtain time - domain waveform data, experimental measurements were conducted on PIN diodes with the following specifications: I - region thickness of \(1~\mu m\); I - region doping concentration of \(2.5\times10^{14}~cm^{-3}\); N - region thickness of \(124~\mu m\); N - region doping concentration of \(5\times10^{19}~cm^{-3}\); P - region thickness of \(2~\mu m\); P - region with Gaussian - type doping and peak concentration of \(1\times10^{20}~cm^{-3}\).

In this work, high - power microwave voltage pulses were applied to the PIN diode, and the corresponding voltage and current waveforms were measured as training data. A total of 160 sets of voltage and current data with different amplitudes and frequencies were collected. The voltage amplitude ranges from 15V to 100V, and the frequency ranges from 1GHz to 10GHz, with uniform sampling. The dataset was divided into training and test sets in an 8:2 ratio, with one full - cycle waveform from each set used for model training. This paper tests four different models: CTRNN, original NCDE, NODE - RNN (first used for Verilog - A behavioral modeling in this paper), and NCDE - RNN (first proposed in this paper). The ODE solver used is the fourth - order Runge - Kutta method. The normalized root mean square error (NRMSE) is used as the evaluation metric on the test set. The modeling experiments were conducted in the Pytorch framework using the torchdiffeq solver library \cite{ref41}. The performance of different models on the test set is shown in Table 1, with the NRMSE metric magnified by two orders of magnitude.

To specifically demonstrate the modeling fitting effects of different models, Figure 2 shows the comparison between the output waveforms of the four models and the true waveforms, with the corresponding input voltage being a standard sine wave.

\begin{figure*}[t]
	\centering
	\includegraphics[width=0.80\textwidth]{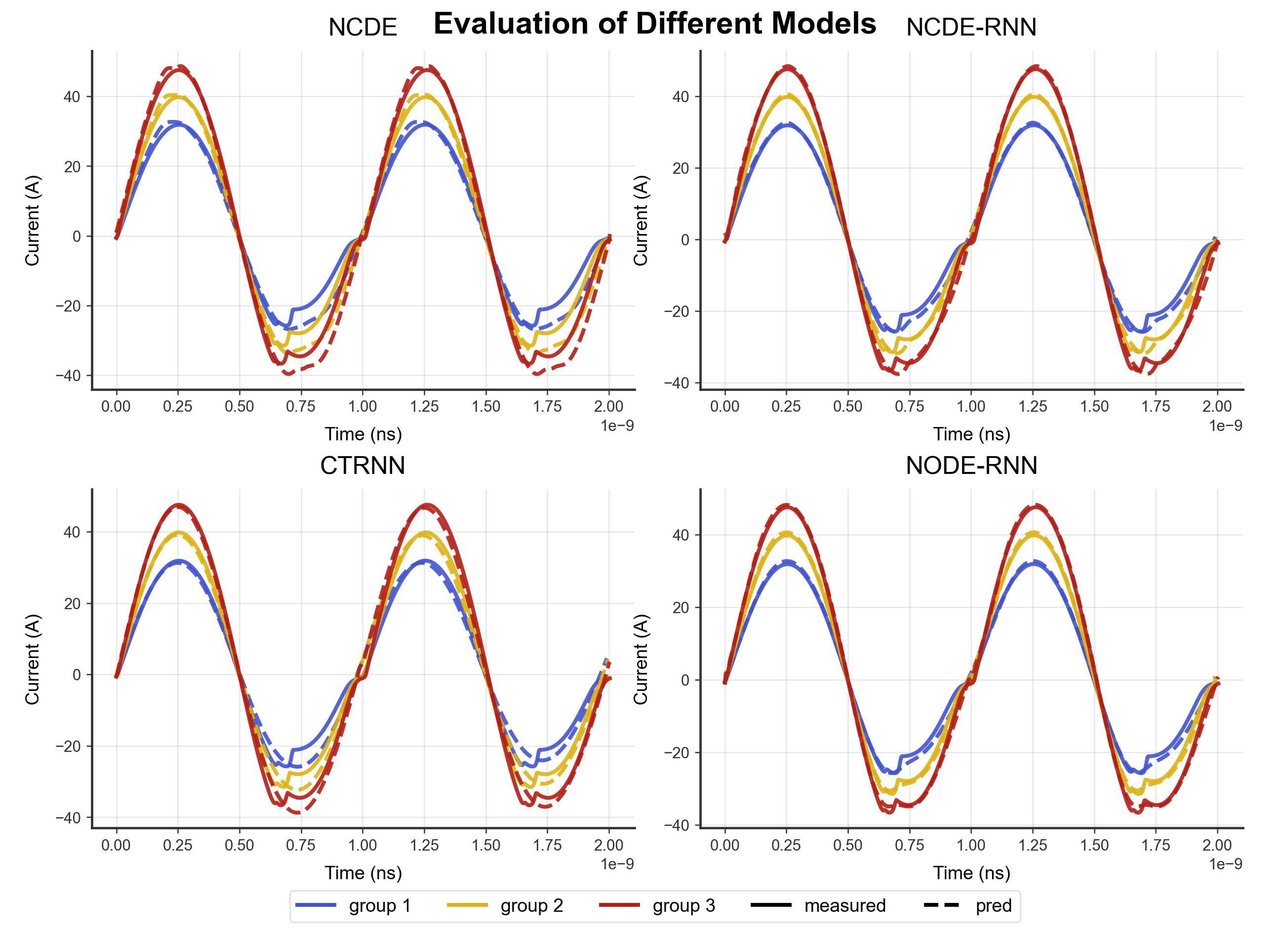}
	\caption{Effect of fitting waveforms of different models}
	\label{fig:screenshot019}
\end{figure*}

\begin{table}
	\centering
	\caption{Evaluation of Different Models on the Test Set}
	\resizebox{\columnwidth}{!}{ 
		\begin{tabular}{|c|c|c|c|}
			\hline
			\textbf{Model} & \textbf{Hidden size} & \textbf{ODESolver} & \textbf{NRMSE}(${10}^{-2}$) \\ \hline
			CTRNN [16] & 27 & RK4 & 4.214 \\  \hline
			NCDE [25]& 16 & RK4 & 4.719 \\  \hline
			NODE-RNN [28]& 16 & RK4 & 3.203 \\  \hline
			\makecell{NCDE-RNN \\ (This work)}  & 16 & RK4 & 3.275 \\  \hline
		\end{tabular}
	}
	\label{tab:example}
\end{table}

\subsection{Model Result Discussion}
Table 1 shows the evaluation metrics of different models on the test set, and Figure 2 illustrates the fitting effects on specific waveforms. It is evident that the hybrid dynamics models significantly enhance modeling capability. Specifically, NCDE-RNN improves fitting effectiveness by 33\% over traditional NCDE, while ODE-RNN enhances performance by 24\% compared to its predecessor CTRNN. As depicted in Figure 5, traditional CTRNN's predicted waveform remains smooth on the negative half-axis, indicating insufficient fitting ability. In contrast, the two new models introduced in this paper can better capture the nonlinear waveforms, showing distinct advantages across different datasets and further demonstrating the superiority of the proposed hybrid dynamics model.

\section{Verification in Verilog-A}
As referenced in \cite{ref16}, the models proposed in this paper can also be deployed in Verilog-A for circuit emulation, as shown in Figure 6. The deployment involves constructing neural network differential equations within circuit equations using internal branches. When Kirchhoff's current and voltage laws are satisfied, the corresponding neural network model is obtained. Notably, the models proposed in this paper require the addition of extra RNN update units.

\begin{figure}
	\centering
	\includegraphics[width=0.99\linewidth]{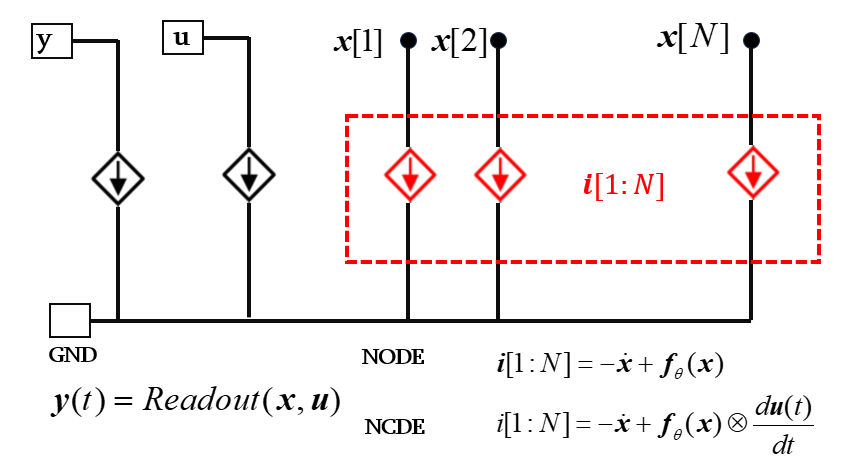}
	\caption{Deployment methods of NODE and NCDE in Verilog-A}
	\label{fig:screenshot020}
\end{figure}

Taking NCDE-RNN as an example, the specific format of the Verilog-A code is as follows.
\begin{lstlisting}[style=myverilog]
`include "constants.vams"
`include "disciplines.vams"
module NCDE-RNN (n, p, gnd);
inout n, p, gnd;
electrical n, p, gnd;
electrical h[0:15];
// Branches for hidden states
branch (h[0], gnd) H0;
...
analog begin
  // Initialization
  if (init_flag == 1) begin
  V(h[0], gnd) <+ 0.0;
  ...
  end
  // get the hiddenstate
  V(h[0],gnd)<+RNN[0];
  ...
  // Based on circuit principles, I(H0) should be 0, establish equation
  //N is hiddenstate size
  I(H0) <+ NN(V(h[0], gnd),...,V(h[N], gnd)) * ddt(V(n, p))/timescale  - time_scale *   ddt(V(h[0], gnd));
  ...
  //RNN cell
  RNN[0]=RNNcell(V(n,p),V(h[0],gnd),...,V(h[N],gnd));
  ...
  //Update the hiddenstate
  ...
  // Readout layer
  readout_sum = Readout_layer(RNN[0], ... RNN[N],V(n,p),ddt(V(n, p));
  I(n, p) <+ readout_sum;
end
endmodule
\end{lstlisting}

\begin{figure}
	\centering
	\includegraphics[width=0.90\linewidth]{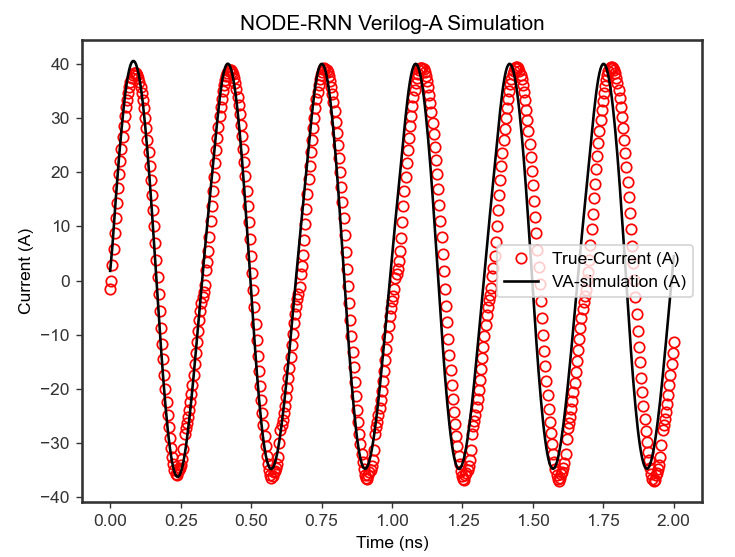}
	\caption{NODE-RNN Verilog-A simulation}
	\label{fig:screenshot017}
\end{figure}

\begin{figure}
	\centering
	\includegraphics[width=0.90\linewidth]{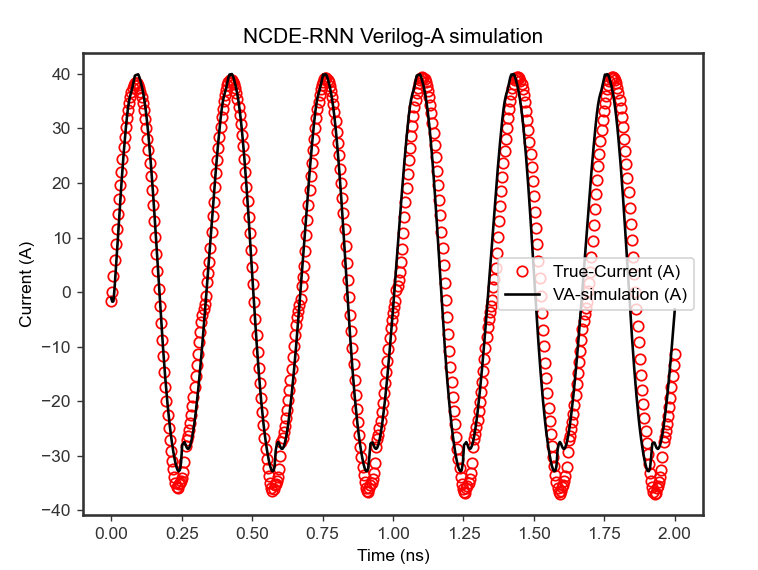}
	\caption{NCDE-RNN Verilog-A simulation}
	\label{fig:screenshot018}
\end{figure}

We deployed the two new models introduced in this paper, NODE-RNN and NCDE-RNN, in Verilog-A to verify their compatibility with current circuit emulation platforms. The comparison between simulation results and real results are shown in Figures 4 and 5, with simulation time and error presented in Table 2. The simulation step is $4e-12$, the simulation time is 2ns, and the input voltage stimulus is a 50V amplitude, 3GHz sine wave. As observed, actual simulation errors are larger than those predicted by neural network models, a point previously noted in \cite{ref16}. This discrepancy arises from circuit simulator settings and solver configurations, which can be optimized to potentially reduce simulation errors.

\begin{table}[h]
	\centering
	\caption{Simulation Evaluation of Different Models in Verilog-A (Time-Related Metrics Under the Same Transient Simulation)}
	\resizebox{\columnwidth}{!}{ 
		\begin{tabular}{|c|c|c|c|}
			\hline
			\makecell{Model \\ } & \makecell{Simulation \\ time (s)} & \makecell{Memory \\ usage (MB)} & \makecell{Simulation \\ NRMSE($10^{-2}$)} \\ 
			\hline
			NODE-RNN & 3.11 & 74.19 & 6.07 \\ \hline
			NCDE-RNN & 3.07 & 77.38 & 7.21 \\ 
			\hline
			
		\end{tabular}
	}
\end{table}

\section{Conclusion}

This paper presents a hybrid dynamics model that combines continuous-time neural differential models with recurrent neural networks (RNNs) for circuit time-domain behavior modeling. It first applies NODE-RNN to Verilog-A modeling and proposes a novel NCDE-RNN model. Experimental results show that this hybrid paradigm significantly boosts nonlinear modeling capability: NCDE-RNN improves fitting effectiveness by 33\% over traditional NCDE models, while NODE-RNN enhances performance by 24\% compared to CTRNN. In high-power microwave tests on PIN diodes, the hybrid models demonstrate a clear advantage in capturing complex nonlinearities and memory effects. Successful deployment and verification in Verilog-A prove the models' compatibility with existing circuit emulation platforms, offering a viable solution for practical engineering applications. This hybrid approach retains the advantages of continuous time-domain modeling and improves response to event-driven dynamic mutations and gradient propagation stability through discrete RNNs. Future work will explore modeling more complex circuits, optimizing model structures to reduce simulation costs, and investigating generalization in other circuit modeling tasks.

\section{Hybrid Adjoint Method}

Training the proposed hybrid models, such as NODE-RNN, requires a specialized backpropagation algorithm capable of handling both continuous-time dynamics and discrete-event updates. Standard adjoint method is not directly applicable. Therefore, we design a hybrid adjoint method that seamlessly integrates continuous adjoint sensitivity analysis for the ODE segments with standard backpropagation for the discrete RNN updates. This approach enables memory-efficient and precise gradient calculation.

\subsection{Model Formulation and Backpropagation Goal}
To derive the method, we first formalize the forward pass of the NODE-RNN model. The process begins with an initial state $\boldsymbol{x}_0$ at time $t_0$. Between any two observation points $(t_{i-1}, t_i)$, the hidden state $\boldsymbol{x}(t)$ evolves according to a neural ordinary differential equation (NODE), where the state at the end of the interval is $\boldsymbol{x}'_i = \text{ODESolve}(\boldsymbol{f}_\theta, \boldsymbol{x}_{i-1}, (t_{i-1}, t_i))$. At each observation time $t_i$, the state is updated by an RNN cell, incorporating the external input $\boldsymbol{u}_i$, resulting in $\boldsymbol{x}_i = \text{RNN}_\phi(\boldsymbol{x}'_i, \boldsymbol{u}_i)$. Finally, the output is computed from the updated state, $\boldsymbol{o}_i = g_{\text{out}}(\boldsymbol{x}_i)$, and the total loss is the sum of losses at each step, $L = \sum_{i=1}^{N} \mathcal{L}(\boldsymbol{o}_i, \boldsymbol{y}_i)$.

Our goal is to compute the gradients of the loss $L$ with respect to the parameters of the continuous dynamics, $\theta$, and the discrete updates, $\phi$.

\subsection{Detailed Derivation of the Hybrid Adjoint Method}
The core of the method is the adjoint state, $\boldsymbol{a}(t)$, defined as the gradient of the total loss $L$ with respect to the hidden state $\boldsymbol{x}(t)$ at time $t$.
\begin{equation}
	\boldsymbol{a}(t) = \frac{\partial L}{\partial \boldsymbol{x}(t)}
\end{equation}
The backpropagation process proceeds backward in time from $t_N$ to $t_0$.

\subsubsection{Initialization}
The process is initialized at the final time step $t_N$. The initial adjoint state is the gradient of the loss with respect to the final hidden state $\boldsymbol{x}_N$.
\begin{equation}
	\boldsymbol{a}_N = \frac{\partial L}{\partial \boldsymbol{x}_N} = \frac{\partial \mathcal{L}(\boldsymbol{o}_N)}{\partial \boldsymbol{o}_N} \frac{\partial g_{\text{out}}(\boldsymbol{x}_N)}{\partial \boldsymbol{x}_N}
\end{equation}

\subsubsection{Backward Iteration from $i=N$ to $1$}
For each time step $i$, the backpropagation involves two main phases: a discrete jump at time $t_i$ and a continuous integration over the interval $(t_{i-1}, t_i)$.

\textbf{Phase 1: Discrete Backpropagation at $t_i$}. At the observation point $t_i$, we backpropagate through the discrete RNN update. Using the chain rule, we compute two quantities.

1) \textit{Adjoint State Propagation:} We compute the gradient with respect to the pre-update state $\boldsymbol{x}'_i$. This gives the initial condition for the subsequent continuous backward pass.
\begin{equation}
	\boldsymbol{a}'_i = \frac{\partial L}{\partial \boldsymbol{x}'_i} = \boldsymbol{a}_i \frac{\partial \text{RNN}_\phi(\boldsymbol{x}'_i, \boldsymbol{u}_i)}{\partial \boldsymbol{x}'_i}
\end{equation}

2) \textit{RNN Parameter Gradient:} We compute the gradient contribution for the RNN parameters $\phi$ at this time step.
\begin{equation}
	\frac{\partial L}{\partial \phi}\bigg|_{t_i} = \boldsymbol{a}_i \frac{\partial \text{RNN}_\phi(\boldsymbol{x}'_i, \boldsymbol{u}_i)}{\partial \phi}
\end{equation}

\textbf{Phase 2: Continuous Backpropagation over $(t_{i-1}, t_i)$}. This is the core of the adjoint method for the continuous part. The dynamics of the adjoint state $\boldsymbol{a}(t)$ are governed by another ODE derived from the forward dynamics.
\begin{equation}
	\frac{d\boldsymbol{a}(t)}{dt} = -\boldsymbol{a}(t) \frac{\partial \boldsymbol{f}_\theta(\boldsymbol{x}(t), t)}{\partial \boldsymbol{x}}
\end{equation}
The negative sign indicates that this system evolves backward in time. The gradient of the loss with respect to the ODE parameters $\theta$ over this interval is given by the integral:
\begin{equation}
	\frac{\partial L}{\partial \theta}\bigg|_{(t_{i-1}, t_i)} = \int_{t_i}^{t_{i-1}} \boldsymbol{a}(t) \frac{\partial \boldsymbol{f}_\theta(\boldsymbol{x}(t), t)}{\partial \theta} dt
\end{equation}
To solve this numerically, we construct an augmented state vector $\boldsymbol{z}(t)$ and solve its dynamics backward in time.
\begin{equation}
	\frac{d}{dt} \begin{bmatrix} \boldsymbol{x}(t) \\ \boldsymbol{a}(t) \\ \frac{\partial L}{\partial \theta} \end{bmatrix} = \begin{bmatrix} \boldsymbol{f}_\theta(\boldsymbol{x}(t), t) \\ -\boldsymbol{a}(t) \frac{\partial \boldsymbol{f}_\theta}{\partial \boldsymbol{x}} \\ -\boldsymbol{a}(t) \frac{\partial \boldsymbol{f}_\theta}{\partial \theta} \end{bmatrix}
\end{equation}
This system is integrated backward from $t_i$ to $t_{i-1}$ with the initial condition $\boldsymbol{z}(t_i) = [\boldsymbol{x}'_i, \boldsymbol{a}'_i, \boldsymbol{0}]$. The forward state $\boldsymbol{x}(t)$ required at each step of the integration is recomputed on-the-fly from $\boldsymbol{x}_{i-1}$ to save memory. The result of this integration yields the adjoint state at the beginning of the interval, $\boldsymbol{a}(t_{i-1})$, and the gradient contribution from this interval.

\textbf{Phase 3: Gradient Accumulation at $t_{i-1}$}. For the next iteration at time $t_{i-1}$ (where $i-1 \ge 1$), the full adjoint state $\boldsymbol{a}_{i-1}$ is the sum of the gradient propagated from the future and the gradient from the local loss.
\begin{equation}
	\boldsymbol{a}_{i-1} = \boldsymbol{a}(t_{i-1}) + \frac{\partial \mathcal{L}(\boldsymbol{o}_{i-1})}{\partial \boldsymbol{o}_{i-1}} \frac{\partial g_{\text{out}}(\boldsymbol{x}_{i-1})}{\partial \boldsymbol{x}_{i-1}}
\end{equation}
This $\boldsymbol{a}_{i-1}$ is then used in Phase 1 of the next backward step.

\subsubsection{Complete Gradient Computation}
After iterating through all time steps from $N$ down to $1$, the total gradients are obtained by summing the contributions from each step.

1) \textit{ODE Parameter Gradient:}
\begin{equation}
	\frac{\partial L}{\partial \theta} = \sum_{i=1}^N \frac{\partial L}{\partial \theta}\bigg|_{(t_{i-1}, t_i)}
\end{equation}

2) \textit{RNN Parameter Gradient:}
\begin{equation}
	\frac{\partial L}{\partial \phi} = \sum_{i=1}^N \frac{\partial L}{\partial \phi}\bigg|_{t_i}
\end{equation}
This hybrid method efficiently computes the exact gradients for the continuous-discrete system without needing to store the entire state trajectory. The derivation for NCDE-RNN follows a similar logic, with the dynamics function $\boldsymbol{f}_\theta$ being replaced by the controlled term $\boldsymbol{f}_\theta(\boldsymbol{x}(t))\frac{d\boldsymbol{U}(t)}{dt}$.

\vfill


\begin{thebibliography}{1}
\bibliographystyle{IEEEtran}

\bibitem{ref1}
T. Liu, S. Boumaiza, and F. M. Ghannouchi, “Dynamic behavioral modeling of 3G power amplifiers using real-valued time-delay neural networks,” {\it{IEEE Transactions on Microwave Theory and Techniques}}, vol. 52, no. 3, pp. 1025–1033, Mar. 2004.

\bibitem{ref2}
D. Luongvinh and Y. Kwon, “Behavioral modeling of power amplifiers using fully recurrent neural networks,” in {\it{IEEE MTT-S International Microwave Symposium Digest}}, Jun. 2005, pp. 1979–1982.

\bibitem{ref3}
Y. Cao and Q.-J. Zhang, “A new training approach for robust recurrent neural-network modeling of nonlinear circuits,” {\it{IEEE Transactions on Microwave Theory and Techniques}}, vol. 57, no. 6, pp. 1539–1553, Jun. 2009.

\bibitem{ref4}
M. Rawat, K. Rawat, and F. M. Ghannouchi, “Adaptive digital predistortion of wireless power amplifiers/transmitters using dynamic real-valued focused time-delay line neural networks,” {\it{IEEE Transactions on Microwave Theory and Techniques}}, vol. 58, no. 1, pp. 95–104, Jan. 2010.

\bibitem{ref5}
I. S. Stievano, I. A. Maio, and F. G. Canavero, “Parametric macromodels of digital I/O ports,” {\it{IEEE Transactions on Advanced Packaging}}, vol. 25, no. 2, pp. 255–264, May 2002.

\bibitem{ref6}
Y. Cao, R. Ding, and Q.-J. Zhang, “State-space dynamic neural network technique for high-speed IC applications: Modeling and stability analysis,” {\it{IEEE Transactions on Microwave Theory and Techniques}}, vol. 54, no. 6, pp. 2398–2409, Jun. 2006.

\bibitem{ref7}
M. Moradi, S. A. Sadrossadat, and V. Derhami, “Long short-term memory neural networks for modeling nonlinear electronic components,” {\it{IEEE Transactions on Components, Packaging and Manufacturing Technology}}, vol. 11, no. 5, pp. 840–847, May 2021.

\bibitem{ref8}
Z. Chen, M. Raginsky, and E. Rosenbaum, “Verilog-A compatible recurrent neural network model for transient circuit simulation,” in {\it{Proceedings of IEEE 26th Conference on Electrical Performance of Electronic Packaging and Systems (EPEPS)}}, Oct. 2017, pp. 1–3.

\bibitem{ref9}
K. S. Narendra and K. Parthasarathy, "Identification and control of dynamical systems using neural networks," in IEEE Transactions on Neural Networks, vol. 1, no. 1, pp. 4-27, March 1990, doi: 10.1109/72.80202.


\bibitem{ref10}
J. Wei and J. M. Dolan, "A robust autonomous freeway driving algorithm," 2009 IEEE Intelligent Vehicles Symposium, Xi'an, China, 2009, pp. 1015-1020, doi: 10.1109/IVS.2009.5164420.



\bibitem{ref11}
M. Tian, J. Bell, E. Azad, R. Quaglia and P. Tasker, "A Novel Cardiff Model Coefficients Extraction Process Based on Artificial Neural Network," 2023 IEEE Topical Conference on RF/Microwave Power Amplifiers for Radio and Wireless Applications, Las Vegas, NV, USA, 2023, pp. 1-3, doi: 10.1109/PAWR56957.2023.10046221.

\bibitem{ref12}
D. R. Morgan, Z. Ma, J. Kim, M. G. Zierdt and J. Pastalan, "A Generalized Memory Polynomial Model for Digital Predistortion of RF Power Amplifiers," in IEEE Transactions on Signal Processing, vol. 54, no. 10, pp. 3852-3860, Oct. 2006, doi: 10.1109/TSP.2006.879264



\bibitem{ref13}

T. M. Martín-Guerrero, J. T. Entrambasaguas and C. Camacho-Peñalosa, "Poly-harmonic distortion model extraction in charge-controlled one-port devices," 2017 12th European Microwave Integrated Circuits Conference (EuMIC), Nuremberg, Germany, 2017, pp. 252-255, doi: 10.23919/EuMIC.2017.8230707.


\bibitem{ref14}

Cenedese, M., Axås, J., Bäuerlein, B. et al. Data-driven modeling and prediction of non-linearizable dynamics via spectral submanifolds. Nat Commun 13, 872 (2022). https://doi.org/10.1038/s41467-022-28518-

\bibitem{ref15}

Daniel, T., Casenave, F., Akkari, N. et al. Model order reduction assisted by deep neural networks (ROM-net). Adv. Model. and Simul. in Eng. Sci. 7, 16 (2020). https://doi.org/10.1186/s40323-020-00153-6


\bibitem{ref16}

J. Xiong, A. Yang, M. Raginsky and E. Rosenbaum, "Neural Ordinary Differential Equation Models of Circuits: Capabilities and Pitfalls," in IEEE Transactions on Microwave Theory and Techniques, vol. 70, no. 11, pp. 4869-4884, Nov. 2022, doi: 10.1109/TMTT.2022.3208896. 


\bibitem{ref17}
Open Verilog International, \textit{Verilog-A Language Reference Manual}, 1996. [Online]. Available: \url{https://www.siue.edu/~gengel/ece585WebStuff/OVI_VerilogA.pdf}



\bibitem{ref18}

Z. Rong et al., "Generic Compact Modeling of Emerging Memories With Recurrent NARX Network," in IEEE Electron Device Letters, vol. 44, no. 8, pp. 1272-1275, Aug. 2023, doi: 10.1109/LED.2023.3290681.



\bibitem{ref19}

Z. Rong, L. Zhang and M. Chan, "Generic Memory Modeling with Recurrent Neural Network," 2022 10th International Symposium on Next-Generation Electronics (ISNE), Wuxi, China, 2023, pp. 1-3, doi: 10.1109/ISNE56211.2023.10221590





\bibitem{ref20}

Y. Cao, R. T. Ding and Q. J. Zhang, "A new nonlinear transient modelling technique for high-speed integrated circuit applications based on state-space dynamic neural network," 2004 IEEE MTT-S International Microwave Symposium Digest (IEEE Cat. No.04CH37535), Fort Worth, TX, USA, 2004, pp. 1553-1556 Vol.3, doi: 10.1109/MWSYM.2004.1338875.




\bibitem{ref21}
J. Xiong, A. S. Yang, M. Raginsky and E. Rosenbaum, "Neural Networks for Transient Modeling of Circuits : Invited Paper," 2021 ACM/IEEE 3rd Workshop on Machine Learning for CAD (MLCAD), Raleigh, NC, USA, 2021, pp. 1-7, doi: 10.1109/MLCAD52597.2021.9531153.




\bibitem{ref22}

F. Bonassi, M. Farina, and R. Scattolini, "Stability of discrete-time feed-forward neural networks in NARX configuration," *IFAC-PapersOnLine*, vol. 54, no. 7, pp. 547-552, Jan. 2021. doi: 10.1016/j.ifacol.2021.08.417.


\bibitem{ref28}

W. Liu, Y. Su and L. Zhu, "Nonlinear Device Modeling Based on Dynamic Neural Networks: A Review of Methods," 2021 IEEE 4th International Conference on Electronic Information and Communication Technology (ICEICT), Xi'an, China, 2021, pp. 662-665, doi: 10.1109/ICEICT53123.2021.9531270.




\bibitem{ref29}

S. A. Sadrossadat, P. Gunupudi and Q. -J. Zhang, "Nonlinear Electronic/Photonic Component Modeling Using Adjoint State-Space Dynamic Neural Network Technique," in IEEE Transactions on Components, Packaging and Manufacturing Technology, vol. 5, no. 11, pp. 1679-1693, Nov. 2015, doi: 10.1109/TCPMT.2015.2484284.



\bibitem{ref38}
P. Kidger, J. Morrill, J. Foster, and T. Lyons, "Neural controlled differential equations for irregular time series," in *Proceedings of the 34th International Conference on Neural Information Processing Systems*, Vancouver, BC, Canada, 2020, art. no. 562, pp. 1-12, Curran Associates Inc., ISBN 9781713829546.

\bibitem{ref39}

 T. Q. Chen, Y. Rubanova, J. Bettencourt, and D. K. Duvenaud, "Neural ordinary differential equations," in Proc. Adv. Neural Inf. Process. Syst., 2018, pp. 6571-6583.

\bibitem{ref40}
C. -T. Tung, S. Salahuddin and C. Hu, "Non-Quasi-Static Modeling of Neural Network-Based Transistor Compact Model for Fast Transient, AC, and RF Simulations," in IEEE Electron Device Letters, vol. 45, no. 7, pp. 1277-1280, July 2024, doi: 10.1109/LED.2024.3404404.


\bibitem{ref41}
Rubanova, Y., Chen, R. T. Q.,  Duvenaud, D. (2019). Latent ODEs for irregularly-sampled time series. In Proceedings of the 33rd International Conference on Neural Information Processing Systems (Art. 478, pp. 1-11). Curran Associates Inc. Red Hook, NY, USA


\bibitem{ref42}
Y. Cao, R. T. Ding and Q. J. Zhang, "A new nonlinear transient modelling technique for high-speed integrated circuit applications based on state-space dynamic neural network," 2004 IEEE MTT-S International Microwave Symposium Digest (IEEE Cat. No.04CH37535), Fort Worth, TX, USA, 2004, pp. 1553-1556 Vol.3, doi: 10.1109/MWSYM.2004.1338875.







\bibitem{ref47}
H. Ge, Y. Liang, J. Lei, C. Yuan and Z. Huang, "Neural ODE Model of Power Electronic Converters With Accelerated Computation and High Fidelity," in IEEE Transactions on Circuits and Systems I: Regular Papers, vol. 71, no. 12, pp. 6363-6374, Dec. 2024, doi: 10.1109/TCSI.2024.3460803.

\end{thebibliography}
\end{document}